%% file: ReID.tex
\documentclass[journal]{IEEEtran}

\input{macro}

\begin{document}

\title{DotSCN: Group Re-identification via Domain-Transferred Single and Couple Representation Learning}

\author{Ziling~Huang$^1$, Zheng~Wang$^1$, Chung-Chi~Tsai$^{2}$, Shin'ichi~Satoh$^{1,3}$, Chia-Wen~Lin$^{4,*}$\thanks{* Corresponding author} \\
$^1$ The University of Tokyo \hspace{5mm} $^2$ Qualcomm Technologies, Inc. \\
$^3$ National Institute of Informatics \hspace{5mm} $^4$ National Tsing Hua University \\
\small{huangziling@gapp.nthu.edu.tw; wangz@hal.t.u-tokyo.ac.jp;
chuntsai@qti.qualcomm.com; satoh@nii.ac.jp; cwlin@ee.nthu.edu.tw}
}

\maketitle

\begin{abstract}
Group re-identification (G-ReID) is an important yet less-studied task. Its challenges not only lie in appearance changes of individuals, but also involve group layout and membership changes.  To address these issues, the key task of G-ReID is to learn group representations robust to such changes. Nevertheless, unlike ReID tasks, there still lacks comprehensive publicly available G-ReID datasets, making it difficult to learn effective representations using deep learning models. In this paper, we propose a Domain-Transferred Single and Couple Representation Learning Network (DotSCN). Its merits are two aspects: 1) Owing to the lack of labelled training samples for G-ReID, existing G-ReID methods mainly rely on unsatisfactory hand-crafted features. To gain the power of deep learning models in representation learning, we first treat a group as a collection of multiple individuals and propose transferring the representation of individuals  learned from an existing labeled ReID dataset to a target G-ReID domain without a suitable training dataset. 2) Taking into account the neighborhood relationship in a group, we further propose learning a novel couple representation between two group members, that achieves better discriminative power in G-ReID tasks. In addition, we propose a weight learning method to adaptively fuse the domain-transferred individual and couple representations based on an L-shape prior. Extensive experimental results demonstrate the effectiveness of our approach that significantly outperforms state-of-the-art methods by 11.7\% CMC-1 on the Road Group dataset and by 39.0\% CMC-1 on the DukeMCMT dataset.
\end{abstract}


\IEEEpeerreviewmaketitle

\section{Introduction}
\label{sec:introduction}
\begin{figure}[t]
\centering 
\includegraphics[width=0.5\textwidth]{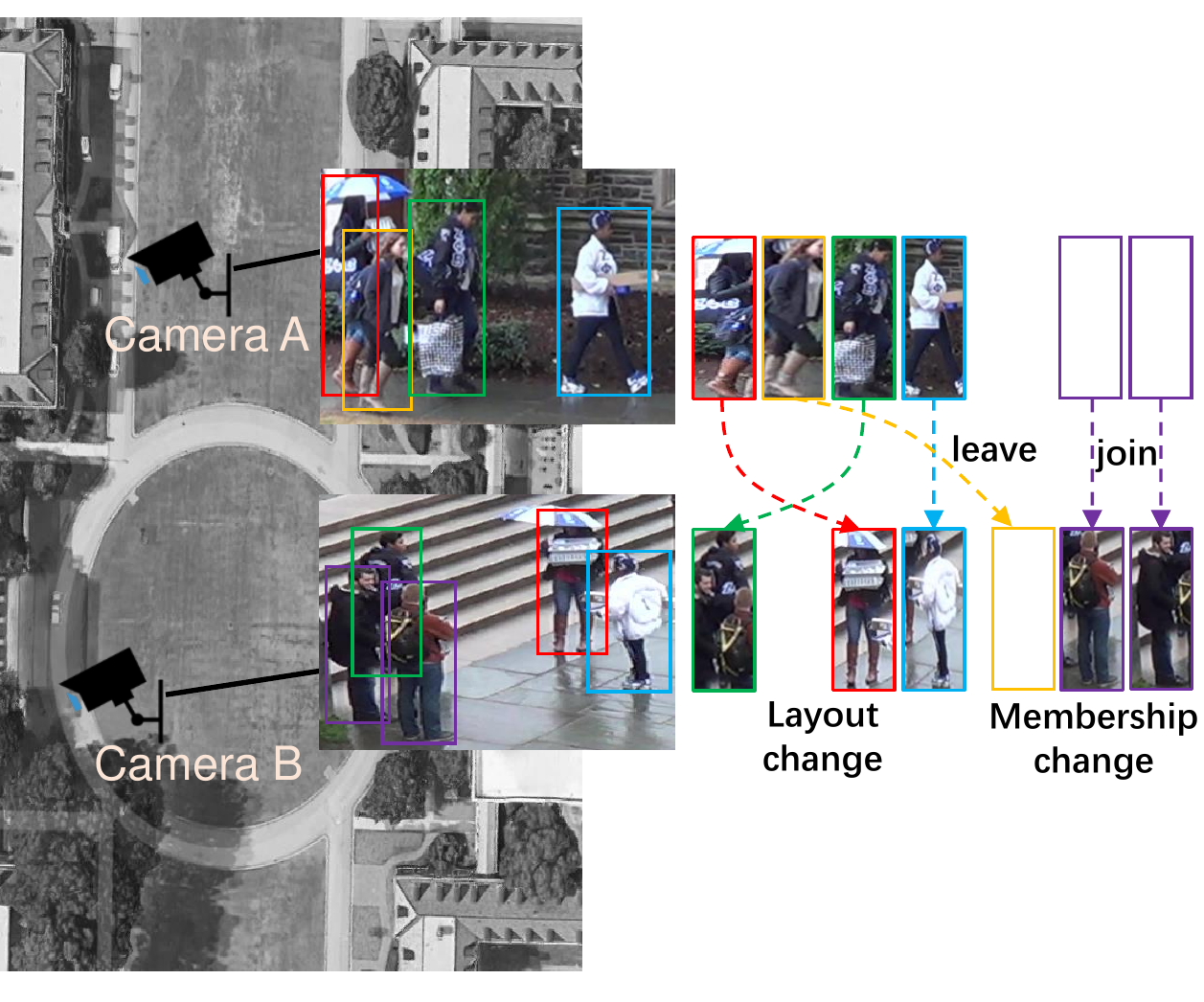}
\caption{{\bf Illustration of the challenges for G-ReID.} Besides the challenge of the appearance change, G-ReID brings in the challenges of group layout and membership changes.}
\label{fig:challenge}
\end{figure}

\IEEEPARstart{D}{ue} to the fast-growing applications in security and surveillance, person/vehicle re-identification (ReID) has been drawing much attention~\cite{li2015cross, wang2016zero, wang2016scale, li2017person, zhong2019poses, zhang2020reid, liu2020reid, huang2020reid, zhu2020dcdlearn}. While existing research works mainly focused on re-identifying individuals, searching out a group of multiple persons simultaneously was relatively rarely studied. In many practical applications identifying a group and then tracking and analyzing the group's behaviors and activities is of fundamental importance. Hence, re-identifying a group of persons (Group ReID or G-ReID) across cameras in different environments is getting more and more important. The G-ReID problem is different from the ReID problem from two perspectives. From the research perspective, G-ReID poses new challenging research problems (will be elaborated in the following) when the ReID target becomes a group of persons. These problems cannot be effectively addressed by existing individual ReID methods as evidenced from the unsatisfactory performances of existing Re-ID methods on G-ReID, and thus call for the development of novel group representations and solutions. From the application perspective, G-ReID is a powerful supplement to individual ReID. For example, criminal cases are often conducted by a certain group of persons rather than an individual. Individual ReID techniques, however, often cannot identify a whole group of suspects in such an application scenario. The technical challenges and practical values of G-ReID have attracted significant amounts of research efforts along this direction.

Different from ReID, G-ReID aims at associating a certain group across different cameras. Besides the challenge of appearance changes for individuals, such as low-resolution, pose variation, illumination variation, and blurred vision, G-ReID raises novel and unique challenges. 1) \textit{Group layout change}. People in a group often change their locations under different camera views, as \figref{challenge} shows. 2) \textit{Group membership change}. Also as illustrated in \figref{challenge}, people will dynamically leave or join a group, thereby making the number of persons in the group change over time. As a result, it is usually not a good choice to treat the group as a whole to extract its global/semi-global features as~\cite{lisanti2017group} did, because the temporal dynamics in a group's layout and membership will alter the visual content of the group. 3) \textit{Lack of labelled training data}. In addition, G-ReID is a new task and lacks annotated image samples with group IDs, \ie, the number of group images is too few to learn robust group representations. We list a comparison of ReID and G-ReID as shown in \tabref{comparison}.

\begin{table}[!htb]
    \centering
    \caption{Comparison of the challenges in ReID and G-ReID}
    \label{tab:comparison}
    \begin{tabular}{L{3.8cm} C{1.9cm} C{1.9cm}}
    \hline
    {\bf Challenge}         &  ReID   &  G-ReID \\
    \hline
    \hline
    {\bf Appearance Change}   &  $\checkmark$  & $\checkmark$   \\
    {\bf Layout Change}       &  $\times$      & $\checkmark$   \\
    {\bf Membership Change}   &  $\times$      & $\checkmark$   \\
    {\bf Training Set} &  Abundant      & No             \\
    \hline
    \end{tabular}
\end{table}

Based on the considerations above, We focus on the two most challenging problems in G-ReID: 1) the problem of  scarce training data, and 2) the problem of layout and membership changes and propose the following new representation learning schemes:

\textbf{Domain-Transferred Single Representations}: Since global features cannot well represent dynamically varying contents in a group, we consider to use single person features to perform group matching as a group image can be treated as a collection of multiple subimages of individual group members. The reason behind is that if we find corresponding persons in a target group, we will search out the target group as well. Considering that training data is difficult to  acquire,~\cite{xiao2018group} exploited hand-crafted features to represent individuals in a group. Nevertheless, hand-crafted representations usually cannot effectively address the appearance change problem in G-ReID due to changes in environments and video capturing conditions. As we know, there exist rich amounts of training datasets suitable for general ReID, which motivates us to make use of existing labeled ReID samples to learn single-person representations. However, the domain gap between the ReID training datasets and target G-ReID images often cause severe performance drop. Therefore, to compensate for the domain shift, we need to find a way to transfer the model learned from existing ReID datasets to better represent new individuals in target G-ReID images. Motivated by the demonstrated success of~\cite{zhong2018generalizing,zhong2018camera}, we propose to transfer the image style of a ReID dataset to that of target G-ReID dataset while preserving individuals' identities. In this way, representative features of individuals in a group can be properly extracted by our transferred representation model.

\textbf{Domain-Transferred Couple Representations}: In G-ReID, we can obtain additional useful information from neighboring group members. Because of group membership changes, it is difficult to exactly determine how many persons are in a group. Regardless of the number of persons in a group, the group can be represented as multiple couple relations. For example, if a group contains three persons $A$, $B$ and $C$, their couple relations expressed as $A-B$, $A-C$ and $B-C$ can be used as effective features for identifying a target group. As a result, we can search out a target group by finding a best-match with the group's corresponding couple relations. Based on this consideration, we propose a couple representation learning network to leverage the information of neighboring individuals. The couple representation learning can also benefit from the proposed domain-transfer technique mentioned above. Note that, in a group, all members can be represented as multiple one-to-one relations (couple representations). Triplet or batch relations can be also represented
by multiple one-to-one relations. That is why we choose the couple representation, which is the simplest unit for social relationship. 

Based on the above discussions, we propose a Domain-Transferred Single and Couple Representation Learning Network (DotSCN), that can offline learn effective features for representing single members and couple relations in a group. We  also propose an online feature fusion method to adaptively combine the two kinds of features together to obtain better group representations. Motivated by the demonstrated success of~\cite{zheng2015query}, our method learns the weights for fusing the single  representations and couple representations without supervision. As will be shown below, our method is easy to implement, yet effective.\footnote{Our code link: \url{https://github.com/huangzilingcv/G-ReID}}. 
Furthermore, the simplicity of our method makes it easy to be applied to different target domains, which is usually not the case for sophisticated methods.

Our contributions lie in the following three aspects:
\begin{itemize}
\item To tackle the scarce training data problem, we  propose a novel method to transfer rich collections of individual ReID samples to the target domain of group ReID to enrich the training samples.
\item We further propose a novel couple representation to capture the social relationship in a group to well address the problem of layout and membership changes for the first time.
\item We propose an online fusion method to adaptively combine the learned single and couple representation results together for better group re-identification.  Extensive experimental results confirm a performance leap from the relevant state-of-the-art techniques in the area.
\end{itemize}

Compared with its preliminary conference version~\cite{huang2019group}, this paper has been significantly expended in several aspects. First, we clarify the differences between ReID and G-ReID that motivates this work and provide a detailed survey of related works. Second, we fully reorganize the offline representation learning scheme and clarify the online adaptive feature fusion scheme based on an L-shape (turning-point) prior. Third, more quantitative tests are conducted in the experiment section along with insightful analyses on the quantitative evaluations. Specifically, we analyze the effectiveness of single and couple features and their fused representation, respectively, and  the complexities of offline representation learning and extraction and online feature fusion.  With these newly added components, the superiority of the proposed method is thoroughly validated.

The rest of this paper is organized as follows.
Some most relevant works are surveyed in Sec. \ref{sec:related}. Sec. \ref{sec:timrl} presents the proposed schemes for transferred representation learning,  couple representation learning, and feature fusion. In Sec. \ref{sec:experiments}, experimental results are demonstrated. Finally, conclusions are drawn in Sec. \ref{sec:conclusion}.

\section{Related Work}
\label{sec:related}

\textbf{Deep learning based ReID}. Deep learning-based approaches have been extensively studied in general ReID field. For example, Li \etal~\cite{li2014deepreid} proposed a filter pairing neural network to jointly handle misalignment and geometric transforms. In order to learn features from multiple domains, Xiao \etal~\cite{xiao2016learning} utilized a domain-guided dropout algorithm to improve the feature learning procedure. Moreover, the method proposed in \cite{su2017pose} makes full use of human part cues to alleviate pose variations and learn robust representations from both a whole image and its different local parts. Chen \etal~\cite{chen2016deep} formulated a unified deep ranking framework that jointly maximizes the strengths of features and metrics. Zhu \etal~\cite{zhu2017part} integrated spatial information for discriminative visual representations by partitioning a pedestrian image into horizontal parts, and proposed a part-based deep hashing network. Yao \etal~\cite{yao2019deep} proposed a part loss network, to minimize both the empirical classification risk on training person images and the representation learning risk on unseen person images. Zheng \etal~\cite{zheng2019pose} introduced  pose-invariant embedding as a pedestrian descriptor and designed a PoseBox fusion CNN architecture. The descriptor is thus defined as a fully connected layer of the network for the retrieval task. However, these supervised learning-based works all require abundant labeled  training data. Moreover, all of these works mainly focused on individual person re-identification. None of them paid attention to G-ReID with very limited training data.

\begin{figure*}[ht]
\centering 
\includegraphics[width=\textwidth]{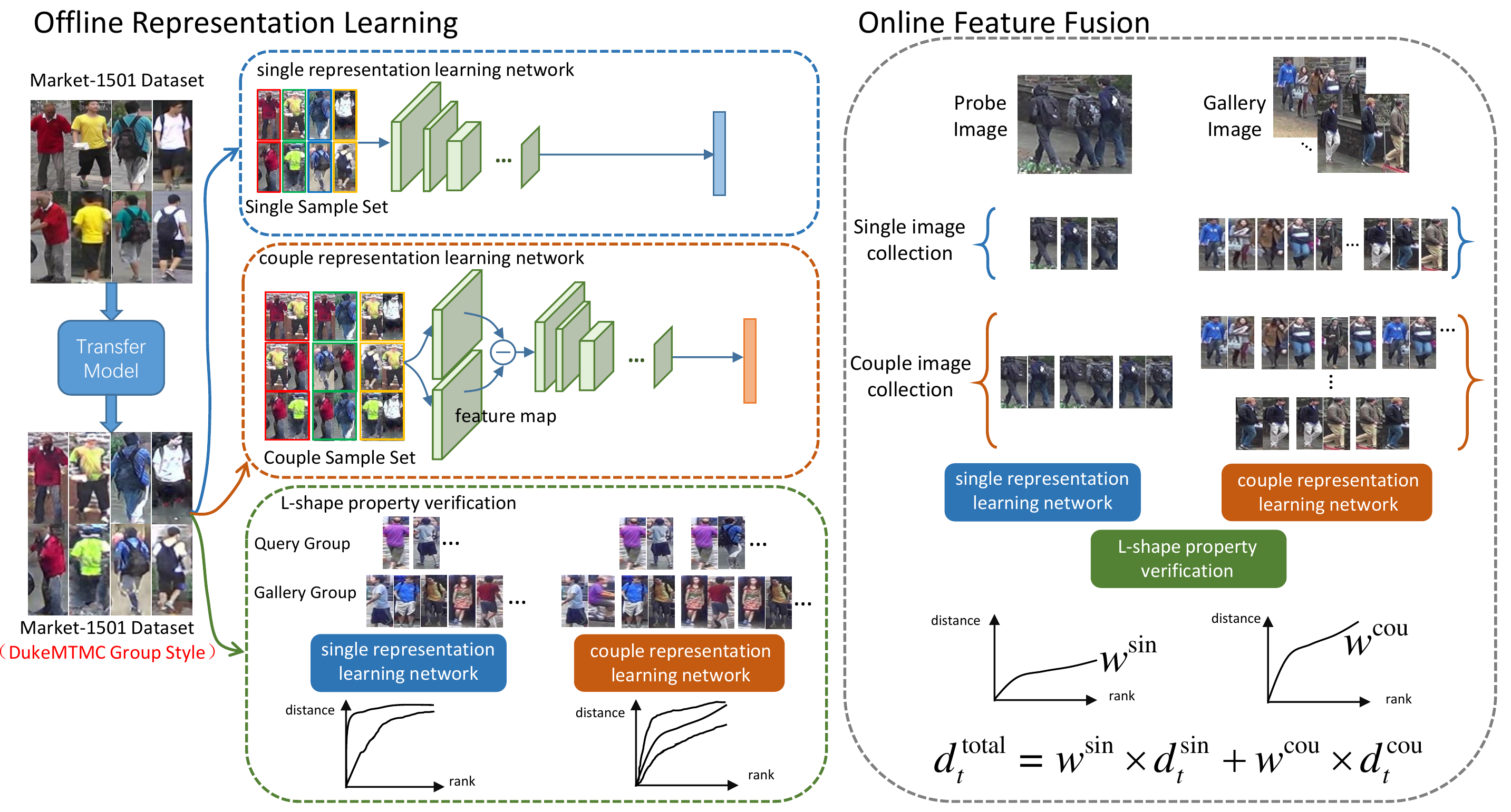}
   \caption{{\bf Proposed architecture.} The proposed Domain-transferred Single and Couple Network (DotSCN) consists of two parts: the offline learning process and the online feature fusion process. Here, we take DukeMTMC group dataset as the example target-domain dataset. In the offline learning process, we first transfer the style of source-domain dataset (\eg, Market-1501) to that of the target one (\eg, DuketMTMC). We construct three sets of transferred training data: the single-person set, couple-person set and group set. We then train DotSCN on the transferred single-person set and couple-person set. The group Set is utilized to verify the L-shape property. In the online identification process, the persons in the group set are cropped to construct the testing single-person and couple-couple sets. Through evaluating the L-shape fitness of the curves, DotSCN learns to fuse the single and couple features to obtain the final representation. Consequently, DotSCN extracts  the single and couple representations of all individual members and couple combinations in a group image, and then measures the distance of the fused single and couple representation  to that of gallery groups.}
\label{fig:framework}
\end{figure*}


\textbf{G-ReID}. Recently, relatively fewer works have focused on G-ReID tasks~\cite{lisanti2017group,xiao2018group, cai2010matching,zheng2009associating,zhu2016consistent,huang2019dot}, compared to general ReID tasks. Some of them mainly attempted to extract global or semi-global features. For example, Cai \etal~\cite{cai2010matching} proposed a discriminative covariance descriptor to obtain both global and statistical features. Zheng \etal~\cite{zheng2009associating} proposed semi-global features by segmenting a group image into several regions. Since persons in a group often change their locations under different views (\ie, layout-change), global and semi-global features are usually sensitive to such changes. In order to make use of individuals' features in the groups, Zhu \etal~\cite{zhu2016consistent} introduced patch matching between two group photos. However, it requires prior restrictions on vertical misalignments, making it unworkable under certain circumstances. Xiao \etal~\cite{xiao2018group} leveraged multi-grain information and attempted to fully capture the characteristics of a group. This approach, however, involves too much redundant information and employs common hand-crafted features, thereby making its accuracy not satisfactory enough. In contrast to the aforementioned methods, our previous work~\cite{huang2019dot} focused on handling the changes on layout and membership of group members using a graph-based group representation. Specifically, it treats a group of individuals as
a graph, where each node denotes the individual feature and each
edge represents the relation between a couple of individuals. Graph samples are then used to train a graph neural network (GNN) to learn a graph group representation for G-ReID.

\textbf{Domain transfer}. Recently, Generative Adversarial Networks (GANs) have been applied to transfer image styles from a source domain to a target domain~\cite{gatys2016image,johnson2016perceptual,bousmalis2017unsupervised,taigman2016unsupervised,deng2018image}. Gatys \etal~\cite{gatys2016image} separated the image content and image style apart and recombined them afterwards, so that the style of one image can be transferred into another. Taigman \etal~\cite{taigman2016unsupervised} proposed a domain transfer network to translate images to another domain while preserving original identities. By making use of the existing domain-transfer techniques, we are able to take advantage of the abundant ReID datasets to improve the performance of our method.

\section{Domain-Transferred Single and Couple Representation Learning}
\label{sec:timrl}

In a G-ReID task, we have a probe image $p$ containing a group of  $N$ persons. 
We aim at finding the corresponding group of probe image $p$ in a set of gallery images $\mathcal{G}=\{g_{t}\}$, where $g_{t}$ represents the $t$-th group image in gallery $\mathcal{G}$. Let $g_{t}^j$ denote the $j$-th person in group image $g_{t}$. As depicted in \figref{framework}, the proposed framework consists of three major parts. First, a domain transfer method is used to transfer the styles of the training ReID dataset into that of the target G-ReID images. In the offline learning process, three sets of transferred training data are respectively  constructed: the single-person set, couple-person set, and group set. We train the proposed DotSCN on the single-person set and couple-person set. The group Set is utilized to verify the L-shape property as will be explained later. In the online identification process, the features extracted by the Single Representation Learning Network (SRLN) and the Couple Representation Learning Network (CRLN) are adaptively fused so that one group can be accurately identified by using the fused single and couple representations. Note that we follow the idea and setting of \cite{xiao2018group} to make the definition of a group in this paper, different from that of \cite{wang2018detecting}. We treat all persons in one image as a ``group''. If two cross-view images from different cameras share more than 60\% members in common, they are considered to contain the same group.

\subsection{Domain Transfer}
Because the total number of people in a collection of G-ReID images is usually rather limited, it is difficult to train a useful network directly based on those available samples. To learn effective representations, we should make use of external information. There exists a rich collection of ReID datasets of individual persons that can help to learn good feature representations. Nevertheless, the domain gap between the existing ReID datasets and the target G-ReID images, caused by their different capturing conditions, can significantly degrade the effectiveness of representation learning. To address this problem, given a training ReID dataset $\mathcal{S}=\{s_{i}\}_{i=1}^{N_s}$, we propose utilizing domain transfer to learn a mapping function $G: \mathcal{S} \rightarrow \mathcal{G}$ from the style of ReID dataset $\mathcal{S}$ to that of G-ReID dataset $\mathcal{G}$ so that the distribution of $G(\mathcal{S})$ can be indistinguishable from that of dataset $\mathcal{G}$. In our work, we exploit the CamStyle\footnote{We refer to the code \url{https://github.com/zhunzhong07/CamStyle}.} method~\cite{zhong2018camera}, to generate the dataset $G(\mathcal{S})$ from the dataset $\mathcal{S}$.

In this way, the dataset $G(\mathcal{S})$, where $y^{s}_{k} \in G(\mathcal{S})$ denotes the $k$-th image of the $s$-th person in the dataset, can be used to train the SRLN $\mathcal{F}^\mathrm{sin}$. In addition, we construct the co-occurrence relations $r^{s_1s_2}=\{(y^{s_1}_{k_1},y^{s_2}_{k_2})|s_1,s_2=1,...,N_s, k_1=1,...,N_{s_{1}}, k_2=1,...,N_{s_{2}}, s_1 \neq s_2\}$ between pairs of persons. These co-occurrence relations are used to train the CRLN $\mathcal{F}^\mathrm{cou}$.

The two networks are then respectively used to extract the single and couple features of gallery images $g_{t}$ and probe image $p$. For probe image $p$, let $\mathcal{F}^\mathrm{sin}(p^i)$ and $\mathcal{F}^\mathrm{cou}(p^{i_1i_2})$ represent its single and couple features, respectively, where $p^{i_1i_2}=\{(p^{i_1},p^{i_2})|i_1,i_2=1,...,N, i_1 \neq i_2\}$ denotes all couple pairs of probe image $p$. For gallery image $g_{t}$, its single and couple features are respectively represented as $\mathcal{F}^\mathrm{sin}(g_{t}^j)$ and $\mathcal{F}^\mathrm{cou}(g_{t}^{j_1j_2})$, where $g_{t}^{j_1j_2}=\{(g^{j_1}_t,g^{j_2}_t)|j_1,j_2=1,...,N_t, j_1 \neq j_2\}$ denotes the couple relations of a gallery image.

\subsection{Offline Representation Learning}
The offline Representation Learning Framework consists of two parts: SRLN for extracting single-person features, and CRLN for extracting joint features between two members in a group. We select ResNet-50 as the backbone CNN structure, since it is the most popular CNN network used in general ReID. Detailed settings about training the network are presented in the experiment section.

To match a group in a collection of group images, the most straightforward method is to find every corresponding people between two groups. We take advantage of an abundance of existing ReID datasets $G(\mathcal{S})=\{y^{s}_{k}\}$ to train the SRLN $\mathcal{F}^\mathrm{sin}$, for which two loss terms are used in the loss function: a cross-entropy loss for classification and a triplet loss for similarity learning.

For G-ReID, besides the personal features of individual group members, the joint features between two co-appearing members are also useful as the co-occurrences of group members have proven to be effective features for characterizing the social relationship in a group~\cite{weng2009rolenet,tsai2013rcnet}. As we know, no matter how many people in an group, they can always be represented as relations between every two group members. In our work, we focus on the relations between two members co-occurring in a group regardless of their spatial distance in the group image as their locations vary in a group dynamically under different camera views.  Based on the above discussion, we propose to represent the couple relations between every two group members by the difference of their personal features to compactly encode the co-occurrence of the two members with only their discrepancy, since the subtraction operation removes their common features. By contrast, the co-occurrence of group members can also be represented by other operations like addition and concatenation. However, adding two person's features magnifies their common features and makes the representations less discriminative, whereas concatenation doubles the number of parameters, thereby significantly increasing the complexity of network training. Specifically, our experimental results show that adopting concatenation for training our model makes it difficult to converge.

Based on the above considerations, the goal of the CRLN $\mathcal{F}^\mathrm{cou}$ is to learn effective occurrence representations of couple pairs of members in a group. To this end, after extracting the features of individual members by the SRLN, the CRLN then pairs all two-member couples in the $G(\mathcal{S})$, and for each couple, subtracts one member's features from the other's to represent their co-occurrence features $r^{s_1s_2}$. Similar to SRLN, a loss function involving a cross-entropy loss term for classification and a triplet loss term for similarity learning is used to train the CRLN. 

\begin{figure} 
    \centering
    \includegraphics[width=\columnwidth]{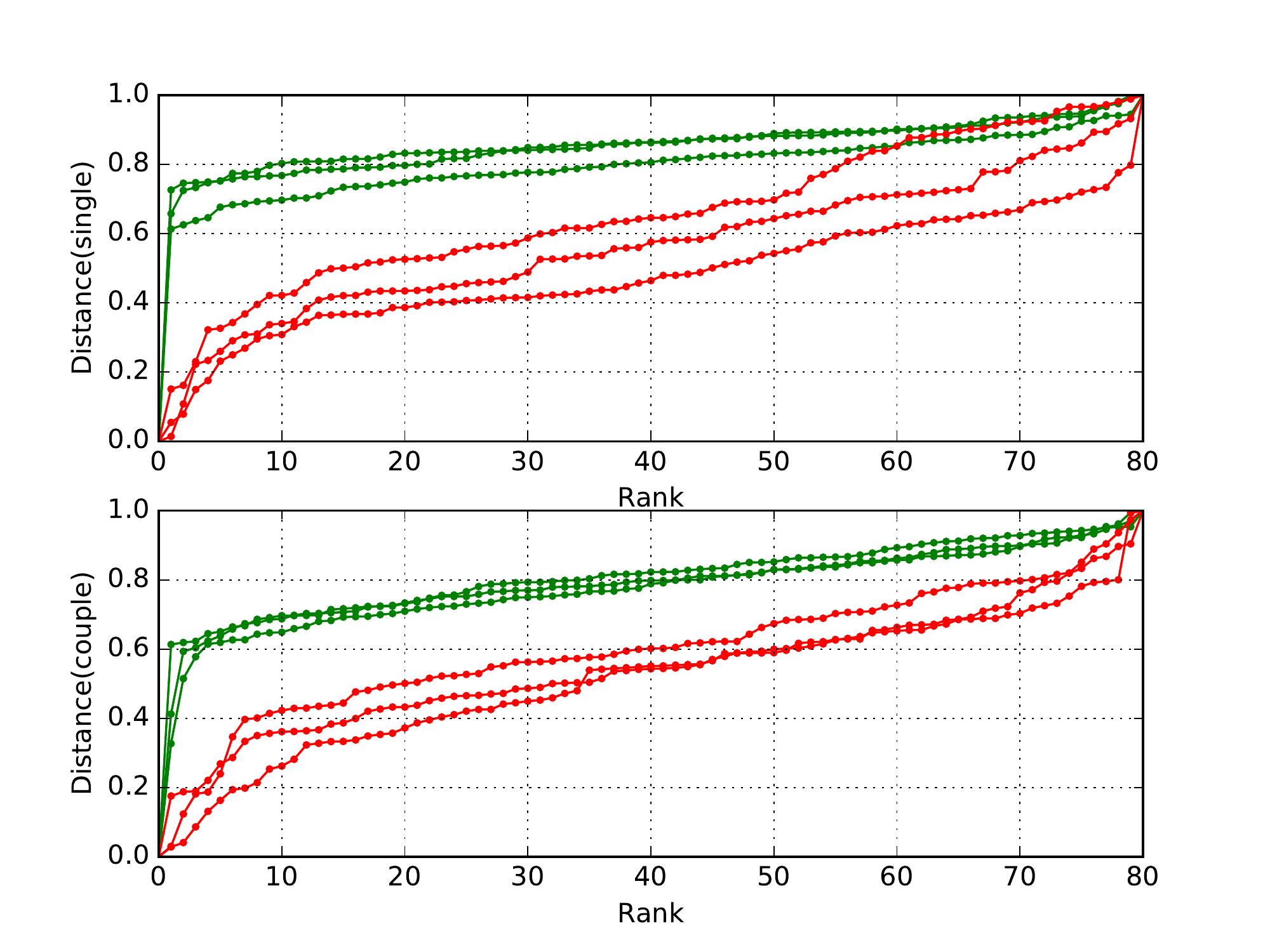}
    \caption{{\bf Logarithmic rank-distance curves for different probes.} Each curve indicates all the distances (sorted in the ascending order) between a probe and all the gallery groups. The top figure shows the results with single features, and the bottom one is generated using couple features. Note that the green lines indicate the representations achieving a good retrieval accuracy, whereas the red lines represent low-accuracy ones. All the high-accuracy curves (\ie, the features are discriminative) are L-shaped, but the property does not hold for the low-accuracy curves.}
    \label{fig:l_shape}
\end{figure}

To utilize the cross-entropy loss to train the CRLN, we assign a unique label to each couple pair of the couple-person set based on a  mapping function $\ell^\mathrm{cou}(i,j)$, where $i$ and $j$ ($i \neq j$) denote the two person labels in a couple-pair. The mapping function $\ell^\mathrm{cou}(i,j)$ should satisfy (i) the commutative property: $\ell^\mathrm{cou}(i,j) = \ell^\mathrm{cou}(j,i)$; and (2) the uniqueness property: $\ell^\mathrm{cou}(i_1,j_1) = \ell^\mathrm{cou}(i_2, j_2)$ if and only if $\{i_1,j_1\} = \{i_2, j_2\}$. 
There are multiple choices of $\ell^\mathrm{cou}$ meeting the above requirements. In our implementation, the couple-pair ID is obtained by
\begin{equation}
    \ell^\mathrm{cou}(i,j)= \frac{1}{2}\max(i,j)^2-\frac{1}{2}\max(i,j)+\min(i,j).
    \label{eq:assign}
\end{equation}

Taking \eqnref{assign} as an example, suppose we have three identities $1$, $2$, $3$ in a group. The couple-pair ID between samples from person $1$ and person $2$ is obtained as $\ell^\mathrm{cou}(1,2) = \frac{1}{2}\cdot2^2-\frac{1}{2}\cdot2 -1 = 0$.
Subsequently, the couple features $r^{s_1s_2}$ along with their unique couple-pair IDs are used to train $\mathcal{F}^\mathrm{cou}$ to learn the  representations of couple pairs.

\subsection{Online Feature Fusion}
\label{subsec:weight_learning}
Since the proposed DotSCN extracts both individuals' features and pairs' features, the two kinds of features need to be adequately fused to obtain better representations for G-ReID. Inspired by \cite{zheng2015query} which shows that given the learned representations for traditional ReID are discriminative, the rank-distance curve between a probe image and all gallery images sorted in the ascending order will exhibits an ``L'' shape, and vice versa. The L-shape property means that there exists a turning point at which the rank-distance curve is significantly flattened out.  This property gives a clue for finding a way of fusing the two kinds of features in DotSCN. As our work is focusing on G-ReID rather than individual ReID,  we need to make sure if the L-shape property still holds with G-ReID features.

To verify the aforementioned property, we use the existing Person-ReID datasets to simulate groups. Specifically, if two images contain the same group of members, they are labeled as the same group. In this step, we have the domain-transferred dataset $\mathcal{G}$ with $N_{s}$ person IDs which are divided into $N_{sub}$ subgroups equally, each containing $N_u=\lfloor\frac{N_{s}}{N_{sub}}\rfloor$ persons in total and being assigned with a unique group ID, where $\lfloor \cdot \rfloor$ means the floor function. Based on the consideration above, we assume that a set of $r_0 \times N_u, \textrm{where } r_0\in [0,1],$ members randomly picked from a group keep staying in the group, whereas the remaining $N_u - r_0 \times N_u$  members may dynamically join or leave the group. No matter how they change their locations or how many people are in the group, they share the same group ID. In this section, we randomly select one combination for each group as a probe image, denoted $p$, the rest of combinations are treated as gallery images, denoted $g_{t}$.

Let $\{\mathcal{F}^\mathrm{sin}(p^{i})\}_{i=1}^{\lfloor\frac{N_{s}}{N_{sub}}\rfloor}$ denote the single features of all individuals in a probe image and $\{\mathcal{F}^\mathrm{sin}(g_{t}^{j})\}_{j=1}^{N_t}$ denote the features of all individuals in the $t$-th gallery image $g_{t}$. The distance between probe image $p$ and gallery image $g_{t}$ is defined as

\begin{equation}
d_{t}^\mathrm{sin}= \frac{1}{N}\sum_{i=1}^{N}~\min{ \{D(\mathcal{F}^\mathrm{sin}(p^{i}),\mathcal{F}^\mathrm{sin}(g_{t}^{j}})| j=1,2,...,N_{t}\},
    \label{eq:score}
\end{equation}
where $D(\cdot)$ denotes the distance metric (say, Euclidean distance in this work). Similarly, we also calculate the couple distance $d_{t}^\mathrm{cou}$ between probe image $p$ and gallery image $g_{t}$ by replacing the single features with the couple features in \eqnref{score}.

The logarithmic rank-distance curves are shown in \figref{l_shape}. We select $\lfloor\frac{N_{s}}{N_{sub}}\rfloor=10$ and $r_0=0.3$ randomly. The green curves indicate the matching distances with discriminative features, while the red ones indicate the matching distances with poor features. Obviously, the green curves are L-shaped,  while the red ones are not. And the green curves enclose larger areas compared with the red ones. Based on the fact above, we propose to fuse the single and couple features to obtain the final distance by applying the following weighted sum:

\begin{equation}
    d_{t}^\mathrm{total}=w^\mathrm{sin} \times d_{t}^\mathrm{sin}+w^\mathrm{cou} \times d_{t}^\mathrm{cou},
    \label{eq:combine}
\end{equation}
where the weights are determined by the enclosed areas of the distance curves with the single and couple features, respectively:
\begin{equation}
    w= e^{\sum_{r}d_t},
    \label{eq:weight}
\end{equation}
where $r$ denotes the rank (\ie, the x-axis of the the curves in \figref{l_shape}). The higher the enclosed area, the more discriminative the features, and the larger the weight.

Based on \eqnref{score}, the  best-match gallery image can be identified by searching for that with the lowest matching distance:

\begin{equation}
    d_p=\min\{d_t^\mathrm{total}|t=1,2,..,N_{t}\}.
    \label{eq:select}
\end{equation}

\subsection{Computational Complexity Analyses}
\label{subsec:time}
The proposed DotSCN consists of two parts. In the offline learning process, the majority of computation is consumed by learning the single and couple representation networks whose complexities depend on the number of training samples. Suppose that there are $N_s$ person images in the training set, the computational complexity of training the single representation network is $O(N_s)$, and that of training the couple representation network is $O(N_s^2)$. Since, in practical applications, we usually extract single and couple features for all gallery images offline in advance, their computation complexities become $O(N_t \cdot N)$ and $O(N_t \cdot N^2)$, respectively, where $N_t$ denotes the number of persons in the gallery image set and $N$ denotes the average number of persons in each image. 

In the online fusion process, the majority of computation is spent on extracting the features of the probe image, calculating the distances between the probe image and gallery images based on the extracted  features, and  deriving the fusion weights. For each probe image, the time cost depends on the amount of probe and gallery features. Suppose that there are $N$ persons in the probe image, the computation complexity will be $O(N\cdot N_t \cdot N + N^2 \cdot N_t \cdot N^2)$ or $O(N^4 \cdot N_t)$. This shows that the online computation complexity counts more on the average number of persons $N$ in each image, implying that the scene complexity of the image dominates the complexity of our algorithm, and the number of gallery images also makes impact.

\section{Experimental Results} \label{sec:experiments}

\begin{figure*}[!htb]
    \centering
    \includegraphics[width=0.8\textwidth]{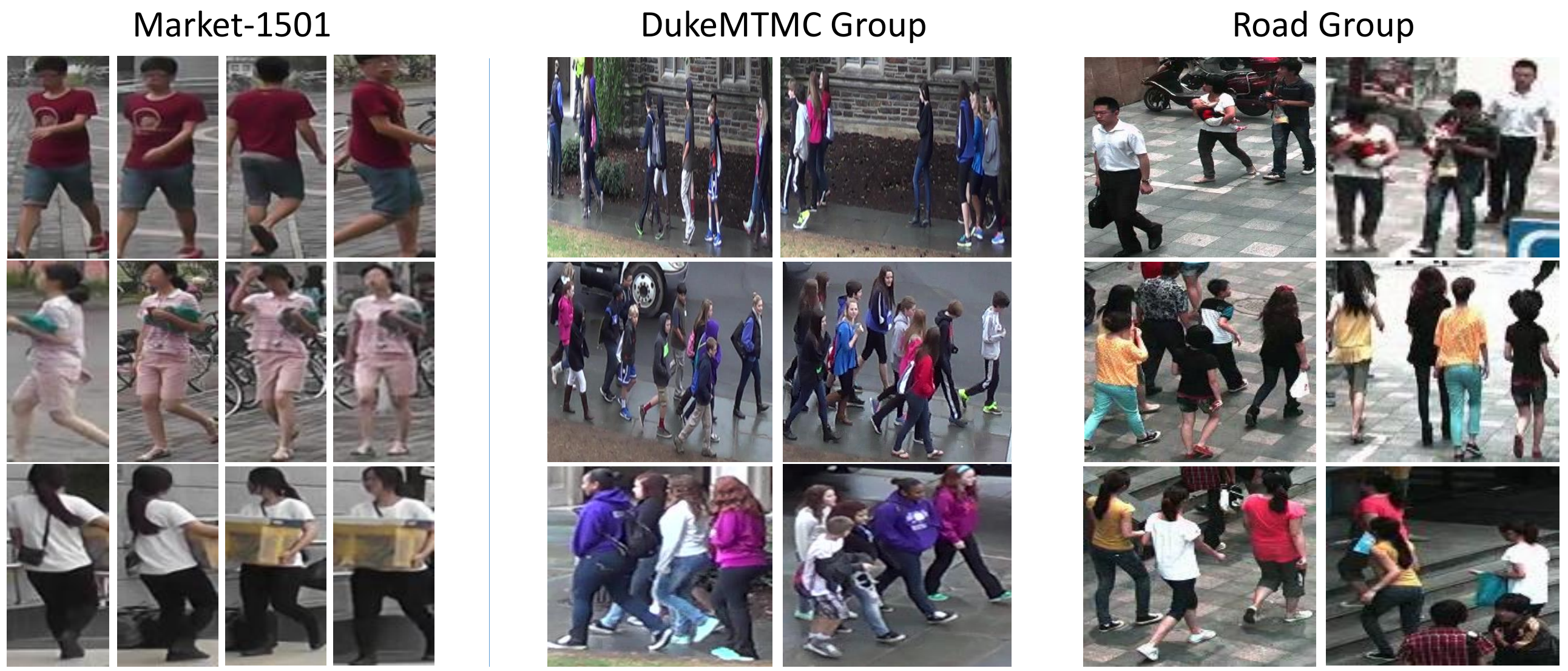}
    \caption{{\bf Snapshots of the utilized datasets.} From left to right, the datasets are respectively \textbf{Market-1501} (ReID), \textbf{DukeMTMC Group} and \textbf{Road Group} (G-ReID). Each row of each dataset shows a few snapshots with the same person/group ID.}
    \label{fig:example}
\end{figure*}

\begin{figure*}[!htb]
    \centering
    \includegraphics[width=0.8\textwidth]{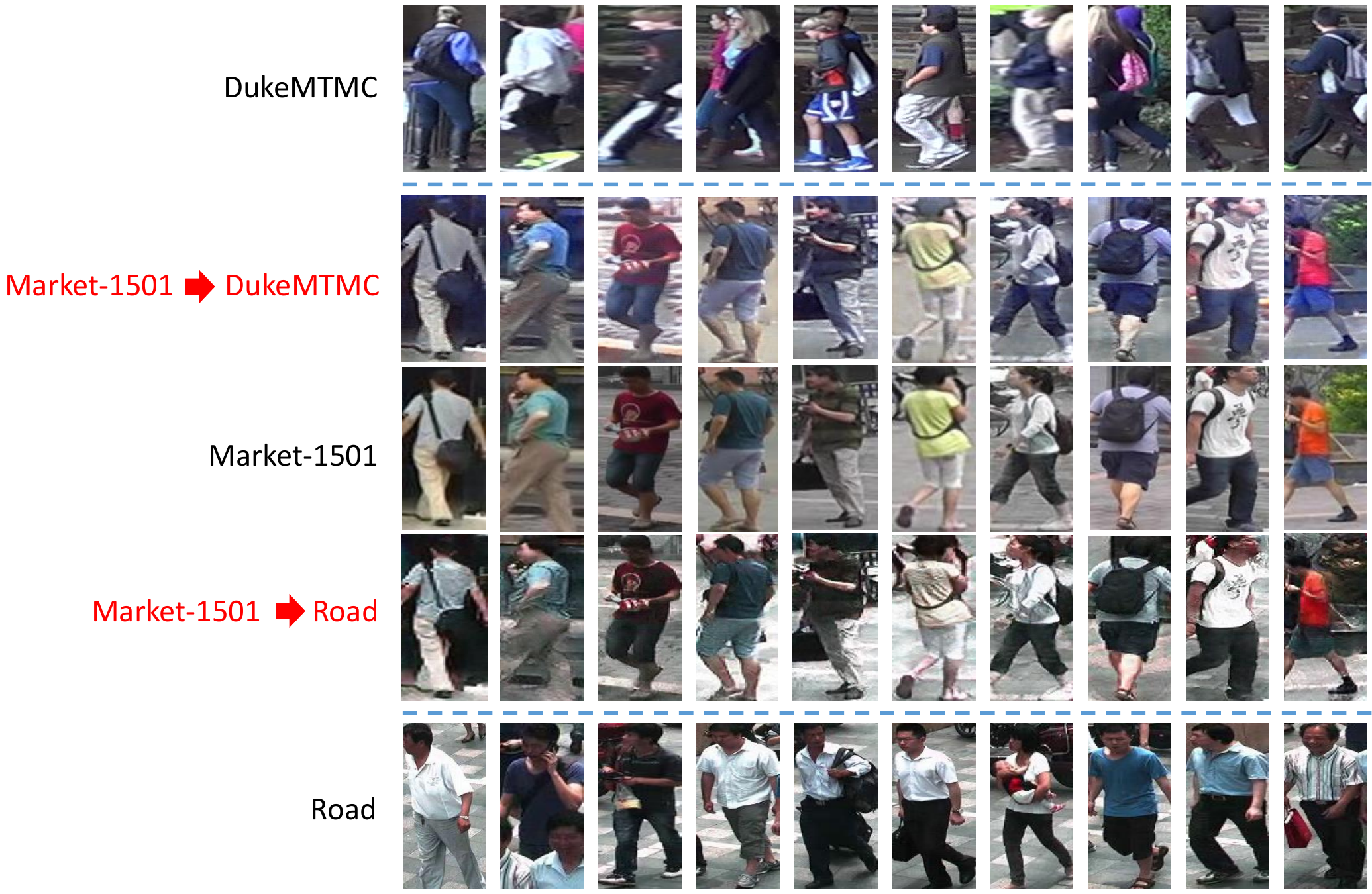}
    \caption{{\bf Snapshots of domain-transferred samples.} The images in the third row are from the source domain \textbf{Market-1501}. The images in the first and fifth rows are cropped respectively from the target domain \textbf{DukeMTMC Group} and \textbf{Road Group}. The second row shows the generated images with \textbf{DukeMTMC} style, and the fourth row shows the generated images with \textbf{Road} style.}
    \label{fig:transfer}
\end{figure*}

\subsection{Datasets and Experiment Setting}
\textbf{Datasets.} Our method is evaluated on two public G-ReID datasets available in~\cite{xiao2018group}\footnotetext{The authors from the universities (in Taiwan and Japan) completed the experiments on the datasets}. Some examples are shown in \figref{example}. The \textbf{DukeMTMC Group} dataset contains 177 group image pairs selected from a 8-camera-view DukeMTMC dataset~\cite{ristani2016performance}. The \textbf{Road Group} dataset contains 162 group pairs taken from a two-camera crowd road scene. These two datasets both include severe object occlusions and large layout \& group membership changes. Following~\cite{xiao2018group}, half of each dataset is evaluated under the protocol in~\cite{zhu2016consistent}, and the Cumulative Matching Characteristic (CMC) metrics~\cite{xiao2018group} are used for performance evaluation. Moreover, we define two cross-view groups as the same one when they have more than $60\%$ members in common.

We also use the Market-1501 dataset~\cite{zheng2015scalable} as the source-domain ReID dataset due to its large amount of training instances: 15,936 images for 751 individuals. In our work, we transfer the domain of Market-1501 dataset to those of \textbf{DukeMTMC Group} and \textbf{Road Group}, respectively, as illustrated in \figref{example}.

\textbf{Setting for Domain Transfer.} In our work, we transfer the domains of existing ReID datasets to that of the target G-ReID datasets (\eg,  \textbf{DukeMTMC Group} and \textbf{Road Group}), prior to training the representations. We use the CycleGAN~\cite{zhu2017unpaired} to transfer the domain for each target G-ReID dataset. As a result, we obtain the DukeMTMC-style Market-1501 dataset and the Road-style Market-1501 dataset. In the training stage, we resize all input images to $256 \times 256$ and use the Adam optimizer. The batch size is 10, and the learning rates are 0.0002 and 0.0001, respectively, for the Generator and the Discriminator.

\begin{table*}[!htb]
    \centering
    \caption{Performance evaluation of the domain-transferred representation learning on \textbf{DukeMTMC Group} and \textbf{Road Group}. The suffix `s' indicates the results with single representations, and the suffix `c' indicates the results with couple representations.}

\begin{tabular}{L{3.4cm} | C{1.35cm} C{1.35cm} C{1.35cm} C{1.35cm} | C{1.35cm} C{1.35cm} C{1.35cm} C{1.35cm}}
\hline
\multirow{2}{*}{Method}   & \multicolumn{4}{c|}{DukeMTMC Group} & \multicolumn{4}{c}{Road Group}\\
             & CMC-1 & CMC-5 & CMC-10 & CMC-20 & CMC-1 & CMC-5 & CMC-10 & CMC-20 \\ \hline\hline
MGR        & 47.4  & 68.1  & 77.3  & 87.4    & 72.3  & 90.6  & 94.1   & 97.5 \\ 
\hline
SCN(s) ResNet50  & 75.0 & 83.0  & 89.8  & 94.3 & 80.2 & 90.1 & 92.6 & \bf{98.8} \\ 
DotSCN(s) ResNet50   & \bf{80.7} & \bf{89.8}  & \bf{94.3}  & \bf{96.6} & \bf{82.7} & \bf{93.8} & \bf{96.3} & \bf{98.8} \\ 
SCN(s) ResNet101  & 78.4  & 84.1   &89.8 & 96.6   &82.7   &90.1   &  93.8 & 98.8  \\ 
DotSCN(s) ResNet101   & \bf{88.8} & \bf{97.7}  & \bf{98.9}  & \bf{100.0} & \bf{83.0} & \bf{96.6} & \bf{97.7} & \bf{100.0} \\ \hline
SCN(c) ResNet50  & 50.0 & 80.7  & 85.2  & 95.5 & 50.6 & 74.1 & 80.2 & 87.7 \\
DotSCN(c) ResNet50   & \bf{62.5} & \bf{83.0}  & \bf{89.8}  & \bf{95.5} & \bf{70.4} & \bf{79.0} & \bf{84.0} & \bf{90.1} \\
SCN(c) ResNet101  &53.4   &83.0   &86.4     &94.3   &56.7   &79.0   &81.4   &90.1   \\ DotSCN(c) ResNet101   & \bf{77.7} & \bf{97.7}  & \bf{97.7}  & \bf{98.9} & \bf{74.0} & \bf{80.2} & \bf{85.1} & \bf{93.8} \\
\hline
\end{tabular}
\label{tab:result_transfer}
\end{table*}

\begin{figure*}
    \centering
    \includegraphics[width=0.8\textwidth]{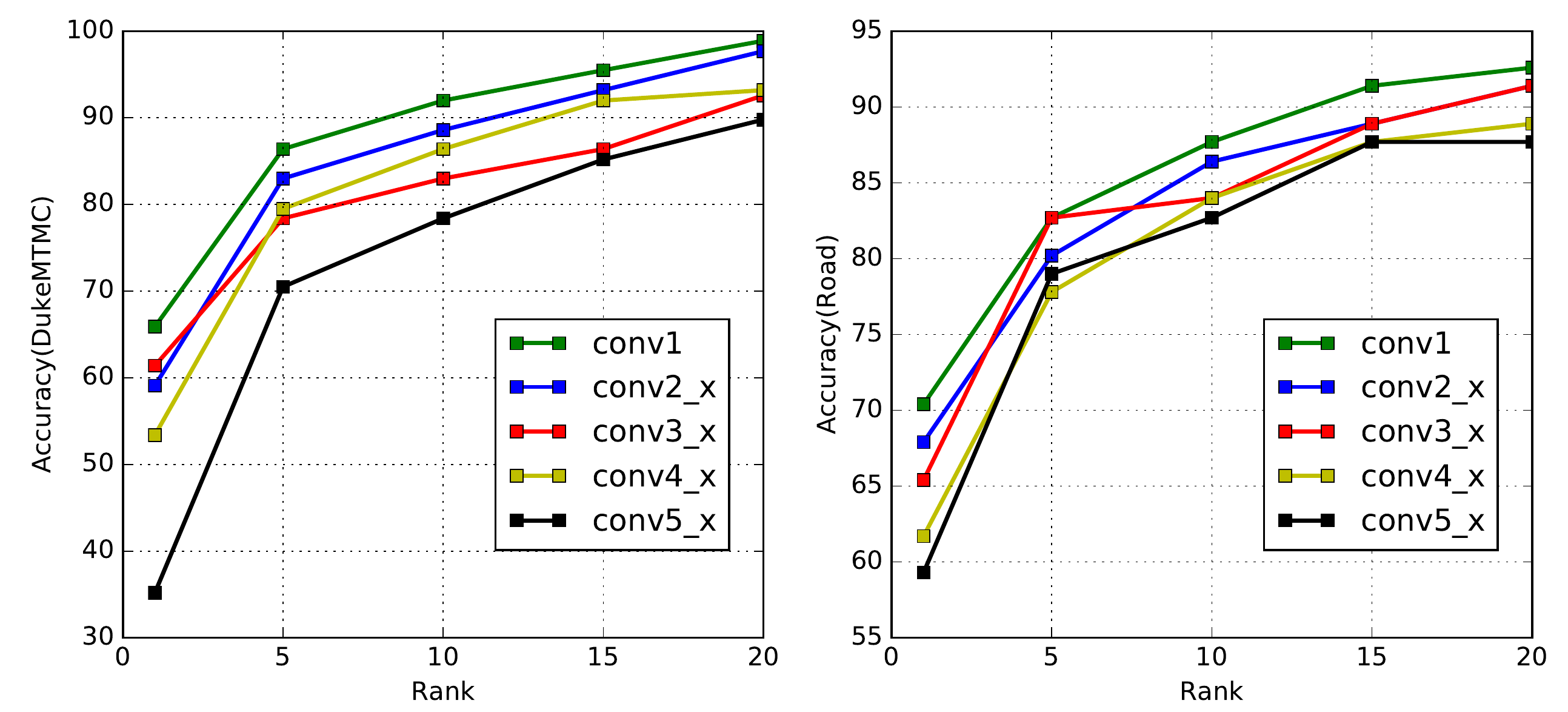}
    \caption{{\bf Effect of the selection of the subtraction layer for couple representation learning.} The curves show the G-ReID accuracy of different network designs of subtraction layers. The two figures respectively show the results (CMC-1) on \textbf{DukeMTMC Group} and \textbf{Road Group}, respectively.}
    \label{fig:l_selection}
\end{figure*}

\textbf{Setting for CNN Models.}
In both SRLN and CRLN, the ResNet-50 with two additional fully connected layers is used as the backbone. The learning rate is set to 0.01, and the dropout rate is set to 0.5. The input images are resized to $256 \times 128$. We use the SGD solver to train the CNN model and set the batch-size to 24. To construct the training set, we pick 100 identities from the domain-transferred dataset, where each identity contains $14-16$ samples. If we use identity $1$ and identity $2$ to form paired samples for a couple class, which contain $C1$ and $C2$ samples respectively, we can generate $C1 \times C2$ couple-pairs for training.

\subsection{Performance of Domain-Transferred Representation Learning}
We conduct experiments to show the effectiveness of domain-transferred representation learning, respectively exploiting the single representations and couple representations learned on \textbf{Market-1501} with/without domain transfer, respectively. For the testing datasets for  G-ReID, we use DukeMTMC/Road samples as the target-domain samples for domain transfer, as illustrated shown in \figref{transfer}.

\tabref{result_transfer} shows the CMC results of the single and couple representations with and without domain-transferred learning, denoted 'SCN' and 'DotSCN', respectively. From the table, we find  that the deep learning features, \ie, `SCN(s)' and `SCN(c)' beat all the hand-crafted feature based schemes in representing a group. We can also observe that the domain-transferred representation learning (\ie, `DotSCN(s)' and `DotSCN(c)') outperforms the non-transferred learning ones. The single representation learning on top of the ResNet50 backbone network achieves 5.7\% and 2.5\% CMC-1 accuracy improvements  on \textbf{DukeMTMC Group} and \textbf{Road Group}, respectively. The couple representation learning achieves 12.5\% and 19.8\% CMC-1 accuracy improvements on \textbf{DukeMTMC Group} and \textbf{Road Group}, respectively. Similarly, our method on top of Resnet101 also achieve significant improvements.

\begin{table*}[!htb]
    \centering
    \caption{Performance evaluation of different couple representations respectively constructed by subtraction and addition on \textbf{DukeMTMC Group} and \textbf{Road Group}. The abbreviation `ADD.' means the couple representations constructed by performing addition on feature maps, and `SUB.' means the couple representations constructed by performing  subtraction on feature maps}
\begin{tabular}{L{3.4cm} | C{1.35cm} C{1.35cm} C{1.35cm} C{1.35cm} | C{1.35cm} C{1.35cm} C{1.35cm} C{1.35cm}}
\hline
\multirow{2}{*}{Method}   & \multicolumn{4}{c|}{DukeMTMC Group} & \multicolumn{4}{c}{Road Group}\\
             & CMC-1 & CMC-5 & CMC-10 & CMC-20 & CMC-1 & CMC-5 & CMC-10 & CMC-20 \\ \hline\hline
DotSCN(c) ADD.   & 44.3 & 56.8 & 80.7  & 86.4  & 48.1 & 70.4 & 79.0 & 88.9 \\
DotSCN(c) SUB.   & \bf{62.5} & \bf{83.0}  & \bf{89.8}  & \bf{95.5} & \bf{70.4} & \bf{79.0} & \bf{84.0} & \bf{90.1} \\
\hline
\end{tabular}
\label{tab:result_add}
\end{table*}

\begin{table*}[!htb]
    \centering
    \caption{Performance evaluation  on different fusion methods on \textbf{DukeMTMC Group} and \textbf{Road Group}. The suffix `s' represents the results with single representations, and the suffix `c' represents the results with couple representations, both extracted by ResNet50 backbone, `EQ.' denotes the fusion results obtained with equal weights, and `AD.' denotes the fusion results obtained with the adaptive weight in \eqnref{combine}}
\begin{tabular}{L{3.4cm} | C{1.35cm} C{1.35cm} C{1.35cm} C{1.35cm} | C{1.35cm} C{1.35cm} C{1.35cm} C{1.35cm}}
\hline
\multirow{2}{*}{Method}   & \multicolumn{4}{c|}{DukeMTMC Group} & \multicolumn{4}{c}{Road Group}\\
             & CMC-1 & CMC-5 & CMC-10 & CMC-20 & CMC-1 & CMC-5 & CMC-10 & CMC-20 \\ \hline\hline
DotSCN(s)    & 80.7 & 89.8  & 94.3  & 96.6 & 82.7 & 93.8 & \bf{96.3} & \bf{98.8} \\ 
DotSCN(c)  & 62.5 & 83.0  & 89.8  & 95.5 & 70.4 & 79.0 & 84.0 & 90.1 \\
DotSCN EQ.         & 80.7 & 89.8  & 94.3  & 96.6 & 82.7 & 90.1 & 92.6 & 95.1 \\ 
DotSCN AD.   & \bf{86.4}  & \bf{98.8}  & \bf{98.8}  & \bf{98.8} & \bf{84.0}  & \bf{95.1}  & \bf{96.3}   & \bf{98.8} \\
\hline
\end{tabular}
\label{tab:result_fusion}
\end{table*}

\begin{table*}[!htb]
    \centering
    \caption{Comparison with the state-of-the-art G-ReID methods on \textbf{DukeMTMC Group} and \textbf{Road Group}}
\begin{tabular}{L{3.4cm} | C{1.35cm} C{1.35cm} C{1.35cm} C{1.35cm} | C{1.35cm} C{1.35cm} C{1.35cm} C{1.35cm}}
\hline
\multirow{2}{*}{Method}   & \multicolumn{4}{c|}{DukeMTMC Group} & \multicolumn{4}{c}{Road Group}\\
             & CMC-1 & CMC-5 & CMC-10 & CMC-20 & CMC-1 & CMC-5 & CMC-10 & CMC-20 \\ \hline\hline
CRRRO-BRO~\cite{zheng2009associating}    & 9.9   & 26.1  & 40.2  & 64.9    & 17.8  & 34.6  & 48.1   & 62.2 \\ 
Covariance~\cite{cai2010matching}   & 21.3  & 43.6  & 60.4  & 78.2    & 38.0    & 61.0    & 73.1   & 82.5 \\ 
PREF~\cite{lisanti2017group}         & 22.3  & 44.3  & 58.5  & 74.4    & 43.0    & 68.7  & 77.9   & 85.2 \\ 
BSC+CM~\cite{zhu2016consistent}      & 23.1  & 44.3  & 56.4  & 70.4    & 58.6  & 80.6  & 87.4   & 92.1 \\ 
MGR~\cite{xiao2018group}        & 47.4  & 68.1  & 77.3  & 87.4    & 72.3  & 90.6  & 94.1   & 97.5 \\
\hline
DotGNN~\cite{huang2019dot} & 53.4  & 72.7  & 80.7  & 88.6 & 74.1  & 90.1  & 92.6   & 98.8 \\
DotSCN (Ours) & \bf{86.4}  & \bf{98.8}  & \bf{98.8}  & \bf{98.8} & \bf{84.0}  & \bf{95.1}  & \bf{96.3}   & \bf{98.8} \\
\hline
\end{tabular}
\label{tab:result_sota}
\end{table*}

\subsection{Performance of Couple Representation Learning}
The CRLN is constructed by the ResNet50 network pre-trained on the domain-transferred data. We divide ResNet50 into six parts based on~\cite{he2016deep}. The difference of the feature maps of different parts can be obtained by subtraction, and is then sent to the rest of the network. The output of the fully-connected layer is taken as the couple representations. The results  are shown in \figref{l_selection}, that shows that if the feature map subtraction is placed closer to the front (conv1), we can obtain better results. That is, the CRLN performs better in representing the relations of low-level feature maps, rather than in the high-level individual features. 

\begin{figure}
    \centering
    \includegraphics[width=\columnwidth]{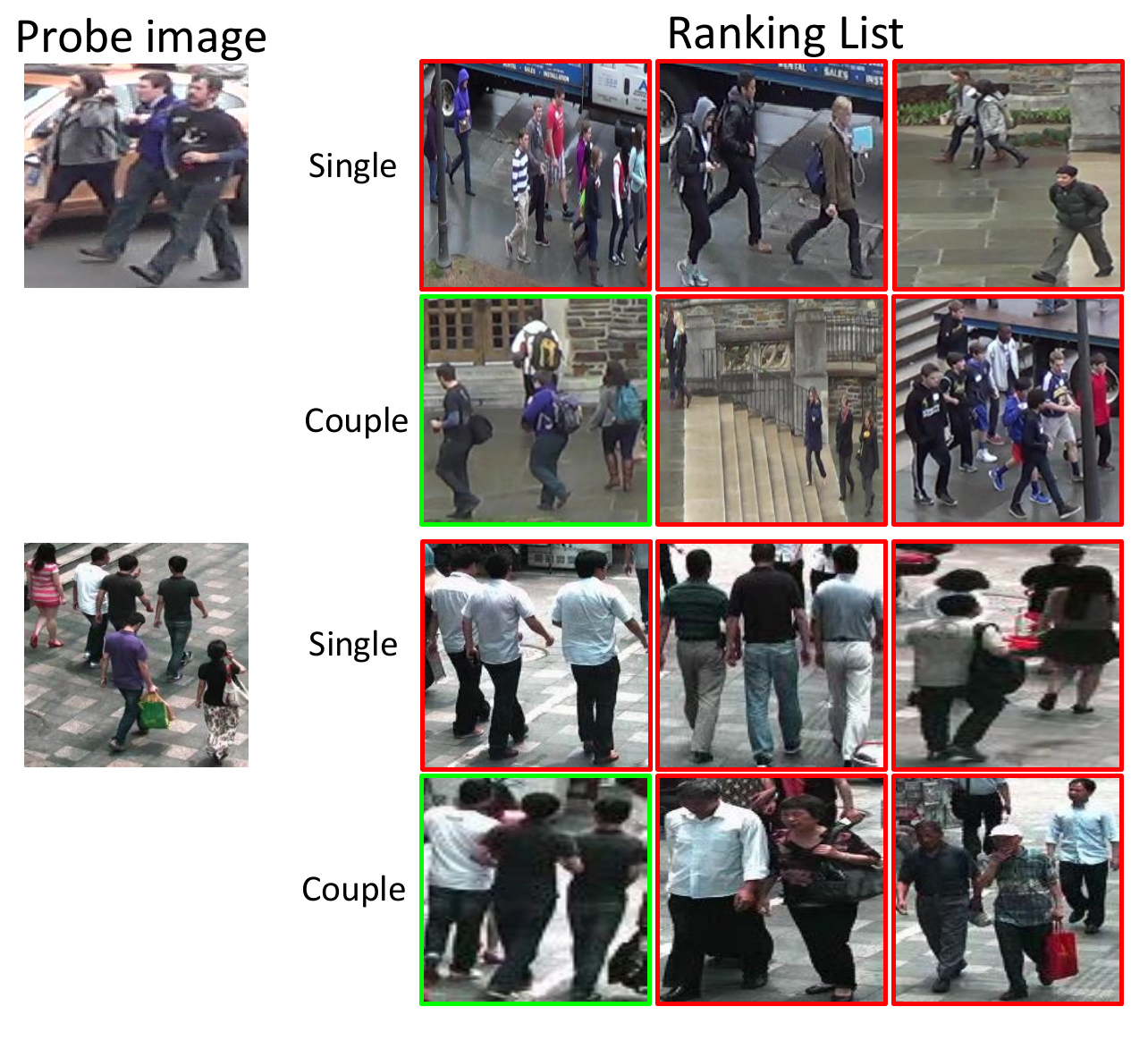}
    \caption{{\bf Two visual examples showing the effectiveness of couple representations.} These two examples are selected from the results of \textbf{DukeMTMC Group} and \textbf{Road Group}, respectively. The first column shows the probe image. The following columns show the top three matches. The samples highlighted by red bounding boxes indicate the negative results, while those highlighted by green bounding boxes indicate the positive results. `Single' and `Couple' represent that the results are obtained respectively by the single or couple representation learning. These two examples show that, in some cases that the single representations lose their effectiveness, the couple representations take their responsibility and offer better predictions.}
   \label{fig:searchresult}
\end{figure}

Moreover, although  \tabref{result_transfer}  shows that the average accuracy with single representations is generally better than that with couple representations, in some cases (e.g., significant appearance variations of individuals due to changes on pose, membership and layout) the couple representations will do a better job and thereby effectively improve the discriminating power via adaptive feature fusion. For example,  \figref{searchresult} illustrates two queries selected from the results of \textbf{DukeMTMC Group} and \textbf{Road Group}, respectively, for which couple representations are more discriminative than single representations. The first column of \figref{searchresult} shows the probe image. The following columns show the top three matches, where the samples highlighted by red bounding boxes indicate the negative results, while those highlighted by green bounding boxes indicate the positive results. The two examples show that, in case that the single representations lose their effectiveness, the couple representations take their responsibility and offer better predictions. Here we also compare the couple representations formed by (1) subtraction operation `DotSCN(c) SUB.' and (2) addition operation `DotSCN(c) ADD.'. \tabref{result_add} shows that the subtraction-based couple representation learning `DotSCN(c) SUB' outperforms the addition-based learning `DotSCN(c) ADD'. 

\tabcolsep=1pt
\begin{figure*}
\centering
\footnotesize{
\begin{tabular}{ccccc}
	$id=14$ &
	\includegraphics[width=0.23\textwidth]{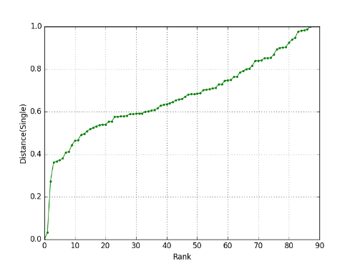} &
	\includegraphics[width=0.23\textwidth]{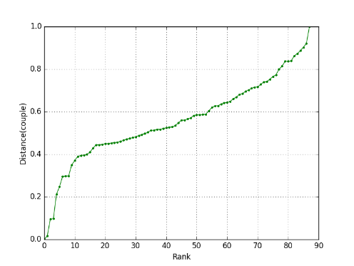} &
	\includegraphics[width=0.20\textwidth,height=2.8cm]{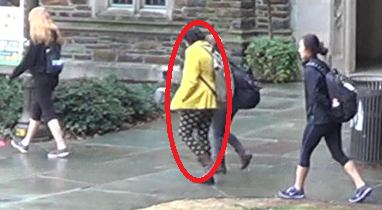} &
	\includegraphics[width=0.20\textwidth,height=2.8cm]{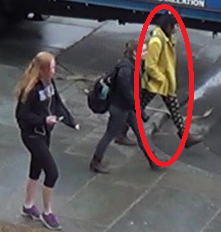}  \\
	$id=41$ &
	\includegraphics[width=0.23\textwidth]{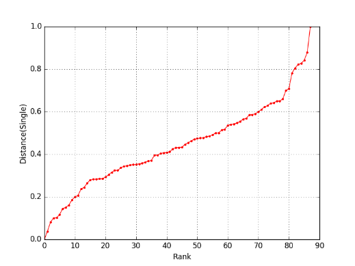} &
	\includegraphics[width=0.23\textwidth]{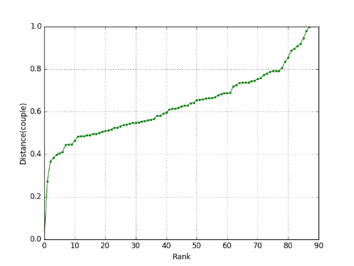} &
	\includegraphics[width=0.20\textwidth,height=2.8cm]{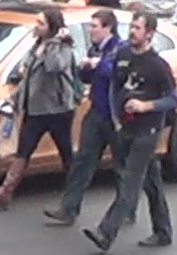} &
	\includegraphics[width=0.20\textwidth,height=2.8cm]{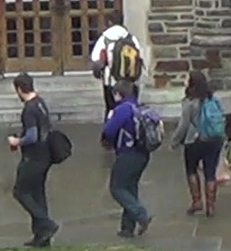}  \\
	$id=10$ &
	\includegraphics[width=0.23\textwidth]{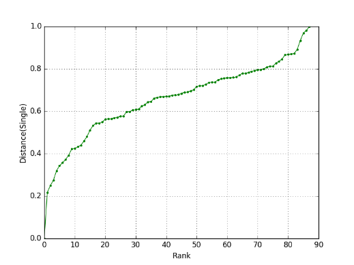} &
	\includegraphics[width=0.23\textwidth]{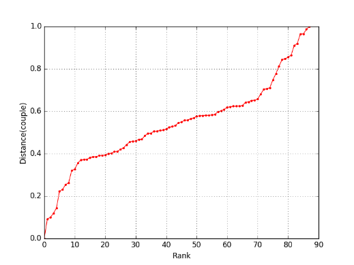} &
	\includegraphics[width=0.20\textwidth,height=2.8cm]{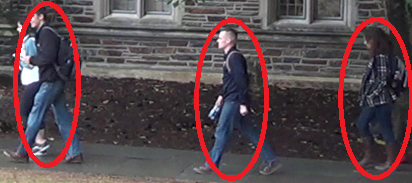} &
	\includegraphics[width=0.20\textwidth,height=2.8cm]{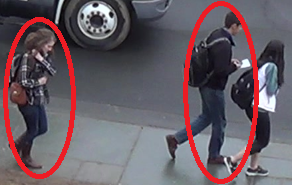}  \\
	& (a) Single Representation & (b) Couple Representation & (c) Camera A  & (d) Camera B \\ 
\end{tabular}}
\caption{Examples showing the selection of single or couple representations. In each row, from left to right, four sub-figures are (a) the logarithmic rank-distance curve generated by the single representations of the given probe image, (b) the logarithmic rank-distance curve generated by the couple representations of the given probe images, (c) the given probe image from camera A, and (d) the matched image of the same group from camera B, respectively. Note that the red curve in (a) indicates that the images (c) and (d) are not perfectly matched by using single representation, whereas the red curve in (b) indicates that the images (c) and (d) are not perfectly matched by using couple representation.}
\label{fig:weight_assign}
\end{figure*}


\subsection{Effectiveness of Adaptive Feature Fusion}
In our method, the final results are generated by fusing results of single and couple representations with \eqnref{combine}. The fusion weights indicate the importance of the representations for the given probe image. To evaluate the proposed adaptive feature fusion `DotSCN AD.' of single and couple representations, we compare it with (1) equal-weight feature fusion `DotSCN EQ.', (2) transferred single representations only `DotSCN(s)', and (3) transferred couple representations only `DotSCN(c)'.  \tabref{result_fusion} shows that the adaptive fusion scheme presented in Section~\ref{subsec:weight_learning} outperforms the other compared methods in all CMC metrics. In contrast, the equal-weight fusion performs the same as transferred single representations on \textbf{DukeMTMC Group}, but slightly worse on \textbf{Road Group}. The results demonstrate the effectiveness of the adaptive weighting scheme guided  by the enclosed areas of the rank-distance curves associated with the single and couple features.

Specifically, to investigate how our method assigns a higher weight to a useful representation, we visualize some typical examples in \figref{weight_assign}. The results of $id=14$ show an example for which both the single and couple representations work well, since one person in the group wears a yellow coat, which is salient making the single and couple representations both discriminative. The results of $id=41$ show an example for which the single representation does not work well, since all the persons in the group wear clothes in very common colors. By contrast, the results of $id=10$ show an example for which the couple representation fails, as the group members wear very similar clothes, making the discrepancy of different persons small. These examples demonstrate that with the proposed online feature fusion scheme, a larger weight will be assigned to the more salient/discriminative representation.

\subsection{Comparison with State-of-the-art Methods}
\tabref{result_sota} compares the performances of our method 
and several state-of-the-art methods on \textbf{DukeMTMC Group} and \textbf{Road Group}. The compared methods include CRRO-BRO~\cite{zheng2009associating}, Covariance~\cite{cai2010matching}, PREF~\cite{lisanti2017group}, BSC+CM~\cite{zhu2016consistent} and MGR~\cite{xiao2018group}. The results show that our method significantly outperforms all existing G-ReID methods, thanks to the proposed transferred representation learning and adaptive feature fusion. In particular, compared with the best state-of-the-art method, our method achieves improvements by 11.7\% CMC-1 on \textbf{Road Group} and 39.0\% CMC-1 on \textbf{DukeMTMC Group}.

To evaluate the effectiveness of the proposed group representation that adaptively fuses single and couple features, we compare it with the graphical group representation model proposed in our previous work~\cite{huang2019dot} that employs a graph neural network (GNN) to represent a group of individuals. The GNN-based framework in \cite{huang2019dot} involves a graph generator for constructing the pool of graph samples with the individuals' representations as nodes and  a GNN model trained on the pool of graph samples for classifying the group IDs. The GNN plays the role of fusing all individuals; features together.  

We implement the 'DotGNN' model that employs the GNN-based group representation model in \cite{huang2019dot} along with the  domain-transfer method used in this work. As demonstrated in \tabref{result_sota}, the adaptive single and couple feature fusion of DotSCN achieves significantly better performance  than the GNN-based group representation of DotGNN in all CMC metrics. Specifically, DotSCN achieves CMC-1 performance improvements by 33.0\% and 9.9\%, respectively, on  \tb{DukeMTMC Group} and \tb{Road Group} compared to DotGNN, because additional multi-person relations (\textit{e.g.}, triplets and 
quadruplets)  other than the couple representation may contain too many similar features, making the GNN difficult to distinguish. 
On the contrary, the couple representation can capture more unique features, effectively complementing the single representation to improve the G-ReID performance.

\section{Conclusion}
\label{sec:conclusion}

In this paper, we addressed an important but rarely studied problem: group re-identification. We have proposed a domain-transferred representation learning and a couple representation learning scheme to respectively overcome the two major challenges with group re-identification: the limited training data challenge and the membership and layout change challenge. We have also proposed an adaptive fusion method to combine the single and couple representations to achieve better group re-identification. Our experimental results have confirmed a performance leap from the relevant state-of-the-art techniques in the area.

\ifCLASSOPTIONcaptionsoff
  \newpage
\fi

\bibliographystyle{IEEEtran}
\bibliography{ReID}

\end{document}

%% file: macro.tex
\usepackage{color}
\usepackage{bbding}
\usepackage{pifont}
\usepackage{wasysym}
\usepackage{amssymb}
\usepackage{amsthm}
\usepackage{booktabs}
\usepackage{multirow} 
\usepackage{setspace}
\usepackage{epsfig}
\usepackage{graphicx}
\usepackage{subfigure}
\usepackage{algorithm}
\usepackage{algpseudocode}
\usepackage{url}
\usepackage{changes}

\newcommand{\etal}{\textit{et~al.}}
\newcommand{\ie}{\textit{i.e.}}
\newcommand{\eg}{\textit{e.g.}}

\newcommand{\figref}[1]{Fig.~\ref{fig:#1}}
\newcommand{\tabref}[1]{Table~\ref{tab:#1}} 
\newcommand{\eqnref}[1]{~(\ref{eq:#1})}

\newcommand{\tb}[1]{\textbf{#1}}

\usepackage{array}
\newcolumntype{L}[1]{>{\raggedright\let\newline\\\arraybackslash\hspace{0pt}}m{#1}}
\newcolumntype{C}[1]{>{\centering\let\newline\\\arraybackslash\hspace{0pt}}m{#1}}
\newcolumntype{R}[1]{>{\raggedleft\let\newline\\\arraybackslash\hspace{0pt}}m{#1}}

\usepackage{cite}
\usepackage{url}
\usepackage[hidelinks]{hyperref}
\hypersetup{
    colorlinks=true,%
    urlcolor=magenta
}